\documentclass[runningheads]{llncs}
\usepackage{graphicx}
\usepackage{amsmath}
\usepackage{multirow}
\usepackage{subfigure}
\usepackage{bbding}
\usepackage{booktabs}
\usepackage[normalem]{ulem}
\useunder{ine}{}{}
\usepackage{bm}
\usepackage{amssymb}
\usepackage{color}
\usepackage{url}
\begin{document}
\title{MACO: A Modality Adversarial and Contrastive Framework for Modality-missing Multi-modal Knowledge Graph Completion}
%
%
\author{
Yichi Zhang,
Zhuo Chen
Wen Zhang\thanks{Corresponding author.}
}
\institute{Zhejiang University, Hangzhou, China     \\\email{\{zhangyichi2022, zhuo.chen, zhang.wen\}@zju.edu.cn}}
%
%
%
\maketitle              
\begin{abstract}

Recent years have seen significant advancements in multi-modal knowledge graph completion (MMKGC). MMKGC enhances knowledge graph completion (KGC) by integrating multi-modal entity information, thereby facilitating the discovery of unobserved triples in the large-scale knowledge graphs (KGs). Nevertheless, existing methods emphasize the design of elegant KGC models to facilitate modality interaction, neglecting the real-life problem of missing modalities in KGs. The missing  modality information impedes modal interaction, consequently undermining the model's performance. In this paper, we propose a modality adversarial and contrastive framework (MACO) to solve the modality-missing problem in MMKGC. MACO trains a generator and discriminator adversarially to generate missing modality features that can be incorporated into the MMKGC model. Meanwhile, we design a cross-modal contrastive loss to improve the performance of the generator. Experiments on public benchmarks with further explorations demonstrate that MACO could achieve state-of-the-art results and serve as a versatile framework to bolster various MMKGC models. Our code and
benchmark data are available at {\color{blue}\url{https://github.com/zjukg/MACO}}.

\keywords{Multi-modal Knowledge Graph  \and Knowledge Graph Completion \and Generative Adversarial Networks.}
\end{abstract}
\section{Introduction}
Knowledge graph completion (KGC) \cite{DBLP:conf/nips/transe} is a popular research topic that focuses on discovering unobserved knowledge in knowledge graphs (KGs) \cite{DBLP:journals/tkde/Survey}, which consist of massive entities and relations in the form of triple \textit{(head entity, relation, tail entity)}. Multi-modal information like images serve as the supplementary information for entities and could also benefit the KGC models, which is known as multi-modal KGC (MMKGC) \cite{DBLP:conf/ijcai/IKRL,DBLP:conf/starsem/TBKGC,DBLP:conf/mm/RSME} in the research community.
\par Typically, MMKGC is accomplished by embedding-based methods, which embed entities and relations in the KGs to a low-dimensional embedding space and design score functions to model the triple structure, thus learning what's known as structural embeddings. Additionally, after feature extraction, multi-modal information such as images needs to be fused and interacted with structural embeddings to improve KGC performance. This highlights the importance of the structural-visual modality interaction and fusion for achieving better MMKGC performance.
\begin{figure}[h]
  \centering
  \includegraphics[width=0.85\linewidth]{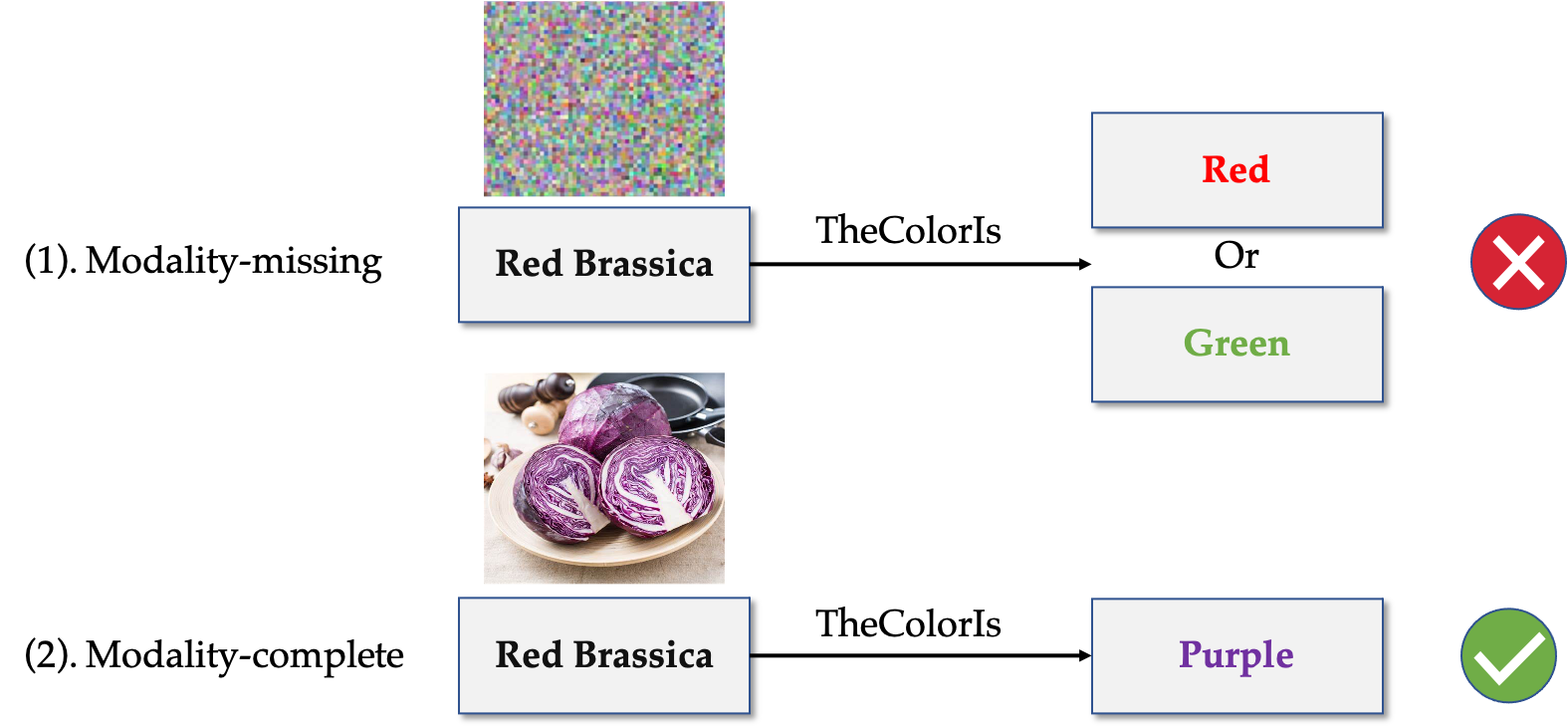}
  
  \caption{A case of the influence of missing modality in KG. Without the help of the visual information, the color of red brassica might be predicted as \textcolor{red}{red} or \textcolor{green}{green} due to the contextual information of KG. The meaningful visual information could guide the KGC model to accurately predict the tail entity.}
  \label{case}
\end{figure}

\par However, construction of real-world KGs typically involves multiple heterogeneous data sources, making it challenging to guarantee complete modality information for all entities and resulting in the modality-missing problem in MMKGC. Such a problem would harm the modality interaction and lead to poor KGC performance. Though existing MMKGC methods \cite{DBLP:conf/ijcai/IKRL,DBLP:conf/starsem/TBKGC,DBLP:conf/mm/RSME} incorporate various approaches to align the structural and visual information, they tend to overlook the modality-missing problem. These methods usually apply simple solutions like random initialization to complete the missing visual information, which might introduce noise into the MMKGC model and loss of some crucial information. Figure \ref{case} illustrates how meaningful visual information could improve the performance of KGC models, which also reflects the importance of completing the visual information of the entity.

\par To address the missing-modality problem, we propose a \textbf{M}odality \textbf{A}dversarial and \textbf{CO}ntrastive (MACO for short) framework for modality-missing MMKGC. Leveraging the generative adversarial framework \cite{DBLP:conf/nips/GAN}, we integrate a pair of generator and discriminator to generate missing visual features conditioned on the entity structural information. Besides, we design a cross-modal contrastive loss \cite{congan2} to enhance the quality of generated features and improve training stability \cite{congan}. The generated visual features would be used in the MMKGC models. To demonstrate the effectiveness of MACO, we conduct comprehensive experiments on the public benchmarks and make further explorations. Experimental results prove that MACO could achieve state-of-the-art (SOTA) KGC results compared with baseline methods and serve as a general enhancement framework for different MMKGC models.
\par The contributions of our work can be summarized as follows:
\begin{enumerate}
    \item We are the first work dedicated to addressing the modality-missing problem in the MMKGC task.
    \item We propose a novel framework MACO to generate realistic visual features and design cross-modal contrastive loss to improve the quality of the generated features.
    \item We demonstrate the effectiveness of MACO with comprehensive experiments on public benchmarks with further exploration, which prove that MACO could achieve SOTA results in the modality-missing MMKGC.
\end{enumerate}

\section{Related Works}
\subsection{Multi-modal Knowledge Graph Completion}
Knowledge graph completion (KGC) aims to discover the unobserved triples in the KGs. Knowledge graph embedding (KGE) \cite{DBLP:journals/tkde/Survey} is a mainstream approach towards KGC. General KGE methods \cite{DBLP:conf/nips/transe,DBLP:journals/corr/distmult,DBLP:journals/jmlr/Complex,DBLP:conf/iclr/rotate} embed the entities and relations of KGs into low-dimensional vector spaces and modeling the triple structure with different score functions.
\par As for multi-modal knowledge graph completion (MMKGC), the modal information (images, textual descriptions) should be considered in the embedding model. IKRL \cite{DBLP:conf/ijcai/IKRL} projects the visual features into the same vector space of structural information and considers the visual features in the score function. TBKGC \cite{DBLP:conf/starsem/TBKGC} further consider visual and textual information and make exploration about modal fusion. TransAE \cite{DBLP:conf/ijcnn/TransAE} employs an auto-encoder to encode the modal information better. RSME \cite{DBLP:conf/mm/RSME} design several gates to select the truly useful modal information. Recent methods like OTKGE \cite{OTKGE} and MoSE \cite{DBLP:conf/emnlp/MOSE} make further steps in multi-modal fusion.
\subsection{Incomplete Multi-modal Learning}
Incomplete multimodal learning (IML) has attracted extensive attention in the research community as the modality-missings situation is common in practice \cite{DBLP:conf/mm/MSA1,DBLP:conf/mm/IML3}. The mainstream solutions towards IML are divided into two categories: the generative methods and the joint learning methods. Generative methods are designed to learn the data distribution and generate the missing modality information with generative frameworks such as GAN \cite{DBLP:conf/nips/GAN} and VAE \cite{DBLP:journals/corr/VAE}. Joint learning methods, however, attempt to learn robust joint embeddings under missing modalities. 

In the KG community, the modality-missing problem has long been neglected. Some multi-modal entity alignment (MMEA) methods \cite{chen2022meaformer,chen2023rethinking} attempt to solve the modality-missing problem.  As for the KGC task, existing methods usually ignore such a problem or just complete the missing information with naive approaches like random initialization. We believe that it is important to complete the missing entity modal information in the process of KGC, to enrich the KGs and improve the performance of KGC.

\section{Methodology}
\subsection{Preliminary}
A KG could be denoted as $\mathcal{G}=(\mathcal{E},\mathcal{R},\mathcal{T})$, where $\mathcal{E},\mathcal{R},\mathcal{T}$ are the entity set, relation set, triple set respectively. As for MMKG, the image set of each entity $e\in\mathcal{E}$ can be denoted as $\mathcal{I}(e)$, which could be $\varnothing$ when the modal information is missing. Furthermore, in the scenario of missing modality, we can partition the entity set into two disjoint parts $\mathcal{E}_{c}$ and $\mathcal{E}_{m}$, which include the modality-complete ($\mathcal{E}_{c}$) and modality-missing ($\mathcal{E}_{m}$) entities respectively.

\subsection{MACO Framework}
\begin{figure}[t]
  \centering
  \includegraphics[width=\linewidth]{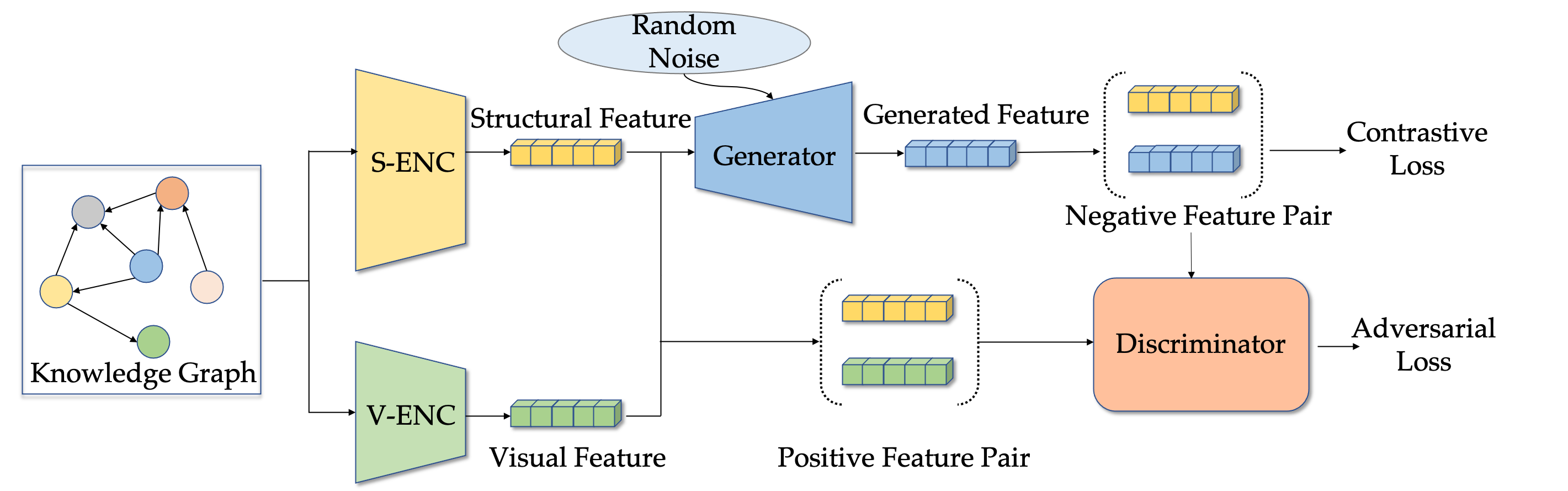}
  \caption{The model architecture of MACO. There are three key designs of MACO: the feature encoders, the adversarial training, and the cross-modal contrastive loss. The structural  encoder (S-ENC) and visual encoder (V-ENC) are used to capture the structural/visual features. The adversarial training would employ a generator and a discriminator and apply adversarial training. The cross-modal contrastive loss is designed to improve the quality of the generated features.}
  \label{model}
\end{figure}

In this section, we will provide a comprehensive overview of our modality adversarial and contrastive framework (MACO) detailedly. A detailed illustration of MACO's model architecture can be found in Figure \ref{model}. MACO is primarily characterized by three key components: feature encoders, modality-adversarial training, and cross-modal contrastive loss. The primary objective of our MACO framework is to complete the visual information of the modality-missing entities.

\subsubsection{Feature Encoders}
We have designed feature encoders to encode the features of different modalities in the knowledge graph (KG). Specifically, we apply a structural encoder $\mathbf{S}$ to encode the structural information of each entity in the KG, while employing a visual encoder $\mathbf{V}$ to encode the visual information of each entity. In our implementation, $\mathbf{S}$ is a $L$-layer relational graph convolution network (R-GCN) \cite{R-GCN}, which could capture the structural features in the KG. For each layer $l (l=1,2,\dots,L)$, the structural features are updated by the message-passing process denoted as:
\begin{equation}
    \bm{s}_{i}^{(l+1)}=\sigma\left(\sum_{r \in \mathcal{R}} \sum_{j \in \mathcal{N}_i^r} \frac{1}{|\mathcal{N}_i^r|} \mathbf{W}_r^{(l)} \bm{s}_{j}^{(l)}+\mathbf{W}_0^{(l)} \bm{s}_{i}^{(l)}\right)
\end{equation}
where $\bm{s}_{i}$ is the structural feature of entity $e_i$, $\mathcal{N}_i^r$ is the neighbor set of $e_i$ under relation $r\in\mathcal{R}$, $\sigma$ is the ReLU activation function \cite{R-GCN}, $\mathbf{W}_0, \mathbf{W}_r$ are the learnable projection matrices.
\par Besides, we employ a vision transformer (ViT) \cite{vit} to capture the visual features of the entities $e_i\in\mathcal{E}_{comp}$. For those entities with more than one image, we apply to mean pooling to aggregate the visual features. The visual feature of entity $e_i$ is denoted as $\bm{v}_i$.
\subsubsection{Modality-adversarial Training}
The second key component of MACO is the modality-adversarial training, which includes a generator $\mathbf{G}$ and a discriminator $\mathbf{D}$. $\mathbf{G}$ is a conditional generator, aiming to generate the visual information given the structural feature of an entity. This design of the conditional generator is intended to enable the generator to produce visual features appropriate for the current entity. Hence, we also term $\mathbf{G}$ the modality-adversarial generator. We implement $\mathcal{G}$ with a two-layer feed-forward network (FFN), which could be denoted as:
\begin{equation}
    \mathbf{G}(\bm{s}, z)=\mathbf{W}_2\left(\delta(\mathbf{W}_1 [\bm{s};z]+\mathbf{b}_1)\right)+\mathbf{b}_2
\end{equation}
where $\mathbf{W}_1,\mathbf{W}_2, \mathbf{b}_1,\mathbf{b}_2$ are the parameters of two feed-forward layers, $\delta$ is the LeakyReLU \cite{Leakyrelu} activation function, $z\sim\mathcal{N}(\bm{0},\mathbf{I})$ is the random noise, and $[;]$ is the concentrate operation. We denote $\bm{g}_i=\mathbf{G}(\bm{s}_i, z)$ as the generated visual feature for entity $e_i$.

\par Moreover, $\mathbf{D}$ serves as a classifier designed to discriminate whether a pair of structural feature $\bm{s}$ and visual features $\bm{v}$ are compatible, which would be a binary classifier. The existing structural-visual feature pair $(\bm{s}_i, \bm{v}_i)$ for  $e_i\in\mathcal{E}_{comp}$ are the positive feature pairs with label 1, while the generated pair $(\bm{s}_i,\mathbf{G}(\bm{s}_i, z))$ are viewed as negative feature pairs with ground-truth label 0.
In practice, $\mathbf{D}$ is another two-layer network denoted as:
\begin{equation}
    \mathbf{D}(\bm{s},\bm{v})=\mathbf{W}_4[\delta(\mathbf{W}_3\bm{s}+\mathbf{b}_3);\bm{v}]+\mathbf{b}_4
\end{equation}
where $\mathbf{W}_3,\mathbf{W}_4, \mathbf{b}_3,\mathbf{b}_4$ are the parameters of the network. 
\par During training, we apply binary cross-entropy as the loss function to optimize the models:
\begin{equation}
    \mathcal{L}_{adv}=-\left(\frac{1}{|\mathcal{E}|}\sum_{e_i\in\mathcal{E}}\log(1-\mathbf{D}(\bm{s}_i,\bm{g}_i))+\frac{1}{|\mathcal{E}_{c}|}\sum_{e_i\in\mathcal{E}_{c}}\log(\mathbf{D}(\bm{s}_i, \bm{v}_i))\right)
\end{equation}
In the adversarial context, the generator $\mathbf{G}$ aims to generate convincing visual features and fool the discriminator $\mathbf{D}$ while $\mathbf{D}$ is designed to make robust predictions to recognize those manually generated features.
Thus, $\mathbf{G}$ and $\mathbf{D}$ would play a mini-max game and optimize their parameters in an adversarial manner, which could be denoted as:
\begin{equation}
\min_{\mathbf{D}}\max_{\mathbf{G}}\mathcal{L}_{adv}
\end{equation}

\subsubsection{Cross-modal Contrastive Loss}
In the mentioned design, we utilized the design concepts of generative adversarial networks (GANs) \cite{DBLP:conf/nips/GAN}, however, the training of GAN models is unstable, and the quality of the generated features is difficult to control\cite{congan2}, potentially decreasing the generator's performance.
\par Thus, we propose another cross-modal contrastive module to contrast the structural features and the generated visual features, aiming to maximize their mutual information and improve the quality of the generated visual features. A pair of structural feature $\bm{s}_i$ and generated visual feature $\bm{g}_i$ of the same entity $e_i$ is regarded as a positve pair and we apply in-batch negative sampling to construct negative pairs. The contrastive loss could be denoted as:

\begin{equation}
    \mathcal{L}_{con}=-\frac{1}{|\mathcal{E}|}\sum_{e_i\in\mathcal{E}}\log\frac{\gamma(\bm{s}_i, \bm{g}_i)}{\gamma(\bm{s}_i, \bm{g}_i)+\sum_{e_j'\in\mathcal{N}(e_i)}\gamma(\bm{s}_i, \bm{g}_j')}
\end{equation}
where $\mathcal{N}(e_i)$ is the negative entity set of $e_i$, $\gamma(\bm{s}_i, \bm{g}_j)$ is the score of a structural-visual feature pair. The score is calculated as:
\begin{equation}
    \gamma(\bm{s}_i, \bm{g}_j)=\exp\left(\mathrm{cos}(\bm{s}_i, \bm{g}_j)/\tau\right)
\end{equation}
where $\mathrm{cos}$ is the cosine similary and $\tau$ is the temperature. In practice, we apply in-batch sampling \cite{congan} to get the negative entities.  When training $\mathbf{G}$, the contrastive loss would be added to the overall objective to enhance the performance of $\mathbf{G}$. Thus, the overall training objective of MACO is:
\begin{equation}
\min_{\mathbf{D}}\max_{\mathbf{G}}\mathcal{L}_{adv} +\min_{\mathbf{G}}\alpha\mathcal{L}_{con}
\end{equation}
where $\alpha$ is the coefficient of the contrastive loss.

\subsection{Missing Modality Completion and Downstream Usage}
Following the above design of MACO, we could obtain the generator $\mathbf{G}$ and a discriminator $\mathbf{D}$. The subsequent step is to complete the missing modality information with $\mathbf{G}$ and $\mathbf{D}$. In our design, for an entity $e_i$, we would first generate $K$ visual features $\bm{g}_i$ by $\mathbf{G}$ and assess their compatibility with the structural feature $\bm{s}_i$ using $\mathbf{D}$. Then we apply to mean pooling to the valid visual feature $\bm{g}_i$ to obtain the final visual feature $\bm{v}_i$. This process can be denoted as $\bm{v}_i=\frac{\sum_{j=1}^Ky_{i, j}\bm{g}_j}{\sum_{j=1}^Ky_{i, j}}$:
where $y_{i, j}\in\{0,1\}$ is the prediction result of $(\bm{s}_i, \bm{g}_j)$ made by $\mathbf{D}$. Further, we propose two strategies to complete the missing modality. The first is to generate only for those modality-missing entities in $\mathcal{E}_m$. The second is to generate for all the entities in $\mathcal{E}$ and change the original visual features for $e\in\mathcal{E}_c$. We name the two strategies as Gen and All-Gen respectively.
\par After generating the visual features, they will be used to initialize the visual embeddings of entities in the KGC model. A score function $\mathcal{S}(h, r, t)$ is designed to measure the triple plausibility, which would calculate the triple score with the structural and visual embeddings. To assign the positive triples with higher scores, we apply margin-rank loss \cite{DBLP:conf/nips/transe} to train the KGC model, denoted as:
\begin{equation}
    \mathcal{L}_{kgc}=\max\left(0,\lambda-\mathcal{S}(h, r, t)+\sum_{i=1}^N p_i\mathcal{S}(h_i', r_i', t_i')\right)
\end{equation}
where $\lambda$ is the margin and $p_i$ is the self-adversarial weight of the negative samples proposed by \cite{DBLP:conf/iclr/rotate}. It is denoted as:
\begin{equation}
    p_i=\frac{\exp(\beta\mathcal{S}(h_i', r_i', t_i'))}{\sum_{j=1}^N \exp(\beta\mathcal{S}(h_j', r_j', t_j'))}
\end{equation}
where $\beta$ is the temperature. During our experiments, we would try several different score functions to demonstrate the effectiveness of MACO.

\section{Experiments}
In this section,  we will present the detailed experiment settings and the experimental results to demonstrate the effectiveness of MACO. We conduct experiments to answer the following three research questions (RQ) about MACO:
\begin{itemize}
    \item \textbf{RQ1}: Could MACO outperform the baseline methods and achieve state-of-the-art results in KGC task?
    \item \textbf{RQ2}: Is the design of each module in MACO reasonable, and is there a pattern to the selection of hyperparameters?
    \item \textbf{RQ3}: Is there a more intuitive explanation for the performance of MACO?
\end{itemize}

\subsection{Experiment Settings}

\subsubsection{Datasets}
For our experiments, we use FB15K-237 \cite{FB15K237} dataset, a public benchmark to conduct our experiments. FB15K-237 has 14541 entities and 237 relations. The train/valid/test set has 272115/17535/20466 triples respectively. The origin FB15K-237 dataset is modality-complete and we construct modality-missing datasets by randomly dropping the visual information of entities with the missing rate (MR) 20\%, 40\%, 60\%, 80\% respectively.

\begin{figure}[h]
  \centering
  \subfigure[IKRL]{\includegraphics[width=\linewidth]{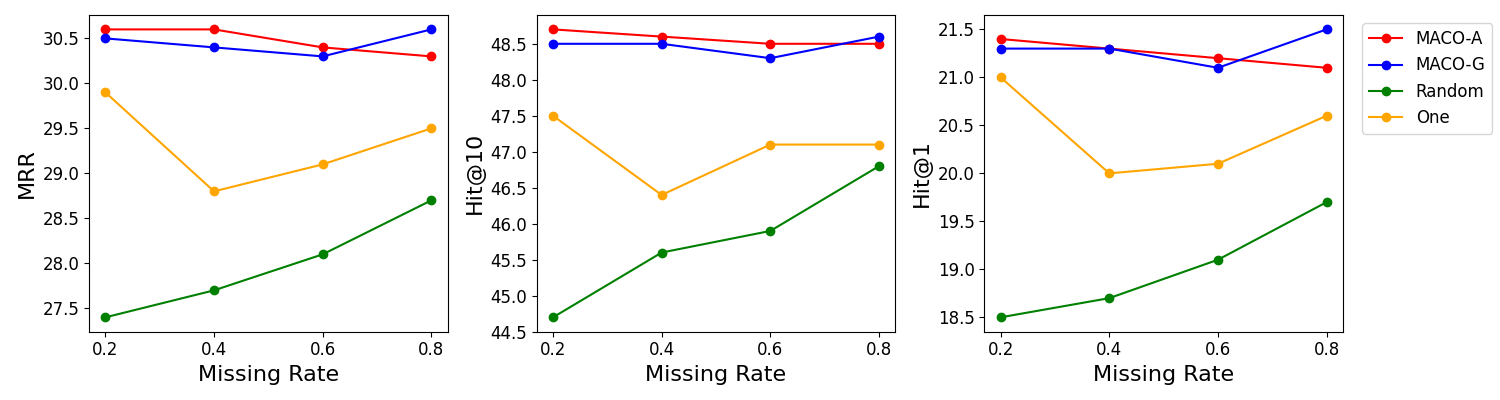}}
  \subfigure[TBKGC]{\includegraphics[width=\linewidth]{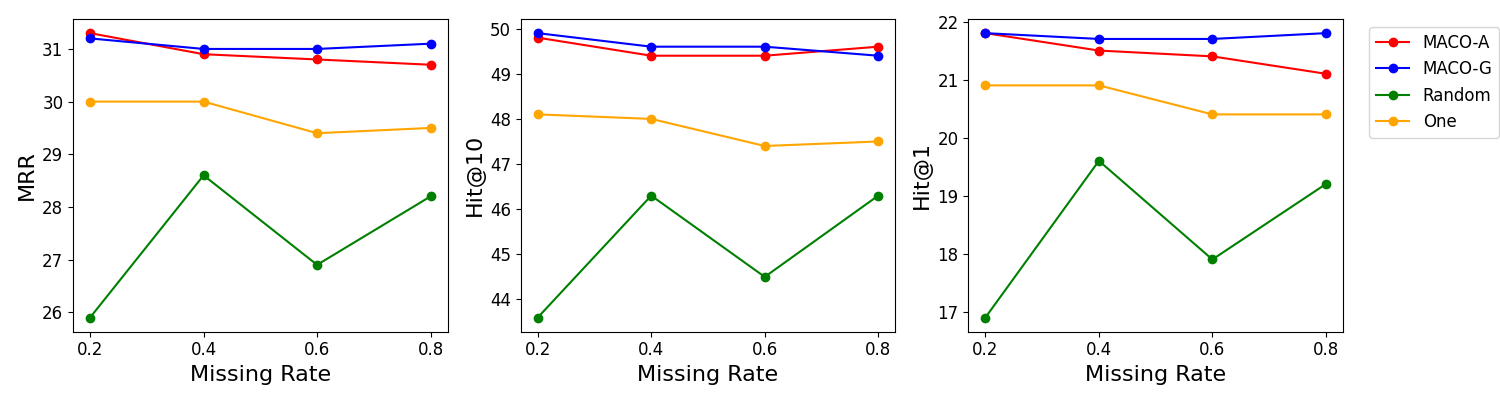}}
  \subfigure[RSME]{\includegraphics[width=\linewidth]{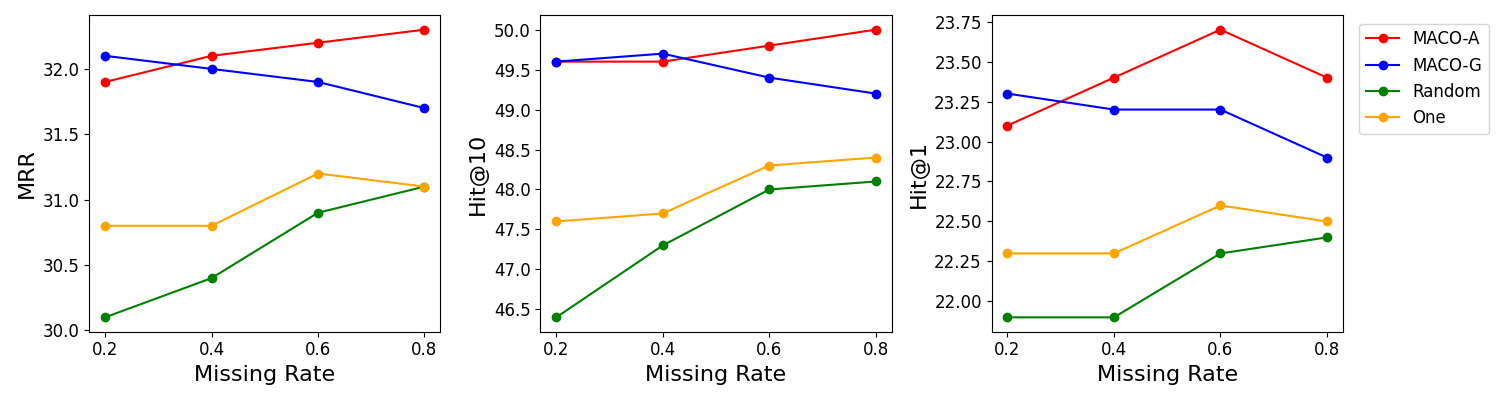}}
  
  \caption{The link prediction results (MRR, Hit@10, Hit@1) compared with baseline methods (Random, One) under different missing rates and different score functions.}
  \label{img::linkprediction}
\end{figure}

\subsubsection{Tasks and Evaluation Protocols}

We evaluate our method with the link prediction task, which is the main task of KGC. The link prediction task aims to predict the missing entities for the given query $(h, r, ?)$ or $(?, r, t)$. We evaluate our method with mean reciprocal rank (MRR), and Hit@K (K=1,3,10) following \cite{DBLP:conf/iclr/rotate}. Besides, we follow the filter setting \cite{DBLP:conf/nips/transe} which would remove the candidate triples that have already appeared in the training data to avoid their interference.

\subsubsection{Baselines}
MACO is designed to complete the missing visual information in the KGs. As few existing works specifically address the modality-missing problem, we have limited choices for baselines. Previous methods often complete the missing modality information by randomly initializing \cite{DBLP:conf/starsem/TBKGC} or setting them all to one \cite{DBLP:conf/mm/RSME}. We name these two methods random and one for short.
Besides, we employ several different score functions (IKRL \cite{DBLP:conf/ijcai/IKRL}, TBKGC \cite{DBLP:conf/starsem/TBKGC}, RSME \cite{DBLP:conf/mm/RSME}) to demonstrate the generality of MACO.

\subsubsection{Parameter Settings}
To train MACO, we set the dimension of structural feature and random noise to 768/128, the number of R-GCN layers $L$ to 2, and the training batch size to 128. The dimension of visual feature captured by ViT \cite{vit} is 768. The hidden size of the FFN is set to 256 for both $\mathbf{G}$ and $\mathbf{D}$. We train MACO for 500 epoches with learning rate $1e^{-4}$ for both $\mathbf{G}$ and $\mathbf{D}$. The temperature $\tau$ is searched in $\{0.5, 1, 2\}$ and $\alpha$ is searched in $\{0.0001, 0.01, 0.1\}$. The number of generated features $K$ is set to 512.

\par As for the link prediction, we fixed the embedding dimension to 128, the batch size to 1024, and the number of negative samples $N$ to 32. The margin $\lambda$ is searched in $\{4, 6, 8\}$ and $\beta$ is set to 2. All experiments are conducted on Nvidia A100 GPUs. Our code and
benchmark data are available at {\color{blue}\url{https://github.com/zjukg/MACO}}.

\subsection{Main Results (RQ1)}
The main results of the link prediction experiments are shown in Figure \ref{img::linkprediction}. From the figures we could observe that MACO could outperform the existing methods on all the evaluation metrics and complete the missing modality with a more semantic-rich representation to achieve better link prediction results under different missing rates with different score functions. Furthermore, we find that the baseline performance is not negatively correlated with the missing rate as expected, but they are significantly lower than MACO, which indicates that vanilla modality completion is not stable for the utilization of modal information.
 
\par Besides, the two strategies (All-Gen and Gen) of MACO exhibit similar performance. They are model-specific as their performance varies across different score functions. For example, All-Gen performs better in RSME, while Gen performs better in TBKGC generally.
Compared to the baseline, the experimental results are more stable under different missing rates, which reflects that MACO could model the distribution of visual information in the graph structure well and generate robust visual representations for entities.

\subsection{Further Analysis (RQ2)}

\subsubsection{Ablation Study}
To answer \textbf{RQ2}, we conduct ablation study and parameter analysis on MACO to demonstrate the effectiveness of each module and hyper-parameters in MACO. In ablation study, we mainly focus on three aspects: (1). the modality-adversarial generator (w/o MA), (2). the R-GCN structural encoder (w/o SE), (3). the contrastive loss function (w/o CL). We remove the mentioned modules respectively and conduct link prediction experiments to explore the quality of the generatedvisual features. Table \ref{ablation} displays the detailed settings and ablation study results, which show that removing any of the modules causes a degradation in results on both score functions. The ablation study indicates that extracting the graph structural features by graph encoder and treating them as the condition of the generator while applying contrastive loss on the features could improve the quality of the generated visual features.
\begin{table}[]
\caption{The ablation study results. We set the missing rate as 40\%. For the model w/o MA, we replace $\mathbf{G}$ with an unconditional generator. For the model w/o SE, we replace the R-GCN encoder with a vanilla embedding layer. For the model w/o CL, we remove the contrastive loss on the training objective.}
\centering
\begin{tabular}{c|cccc|cccc}
\toprule
\multirow{2}{*}{Model} & \multicolumn{4}{c|}{IKRL}     & \multicolumn{4}{c}{RSME}      \\
                       & MRR  & Hit@10 & Hit@3 & Hit@1 & MRR  & Hit@10 & Hit@3 & Hit@1 \\ \midrule 
MACO                   & 30.6 & 48.6   & 34.1  & 21.3  & 32.1 & 49.6   & 35.1  & 23.4  \\
w/o MA                 & 29.6 & 47.5   & 32.8  & 20.4  & 31.4    & 48.5      & 34.4     & 22.8     \\
w/o SE                 & 29.6 & 47.6   & 32.8  & 20.6  & 31.3    & 48.2      & 34.3     & 22.7     \\
w/o CL                 & 29.7 & 47.7   & 32.9  & 20.7  & 31.3 & 48.4   & 34.3  & 22.8 \\
\bottomrule
\end{tabular}
\label{ablation}
\end{table}

\subsubsection{Parameter Analysis}
\begin{figure}[h]
  \centering
  \subfigure[Constrative Temperature]{\includegraphics[width=0.45\linewidth]{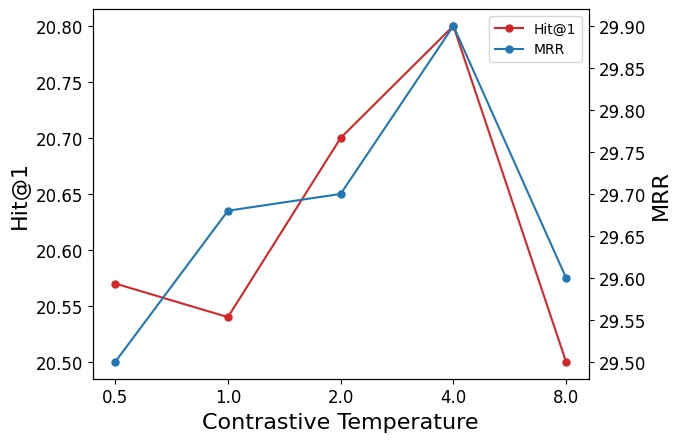}}
  \subfigure[Loss Coefficient]{\includegraphics[width=0.45\linewidth]{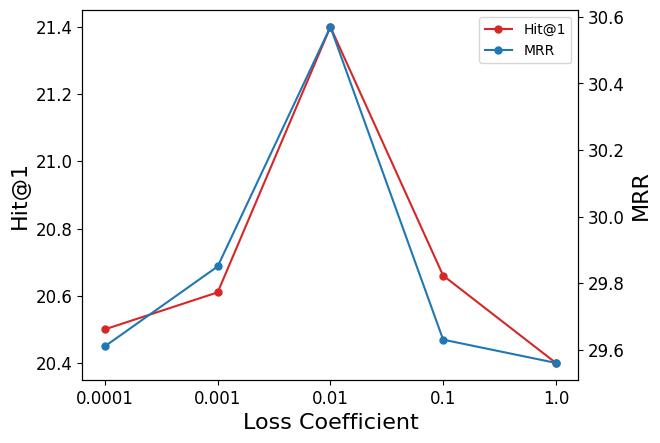}}
  
  \caption{Parameter analysis results of MACO. The missing rate of dataset is  60\% and IKRL \cite{DBLP:conf/ijcai/IKRL} score function is employed for the parameter analysis.}
  \label{parameter}
\end{figure}

We further evaluate the influence of the hyper-parameters of MACO including the temperature $\tau$ and the contrastive loss coefficient $\alpha$, which are newly introduced in MACO. Figure \ref{parameter} reveals that the two hyper-parameters significantly affect the model's performance. Additionally, they show a similar impact on the model performance, with an initial improvement observed as the hyperparameters increase, which is later followed by a performance decrease. Empirically speaking, the optimal choice of $tau$ and $\alpha$ is near 4.0 and 0.01 respectively.

\subsection{Case Study (RQ3)}

\begin{figure}[t]
  \centering
  \subfigure[Modality-missing triples]{\includegraphics[width=0.45\linewidth]{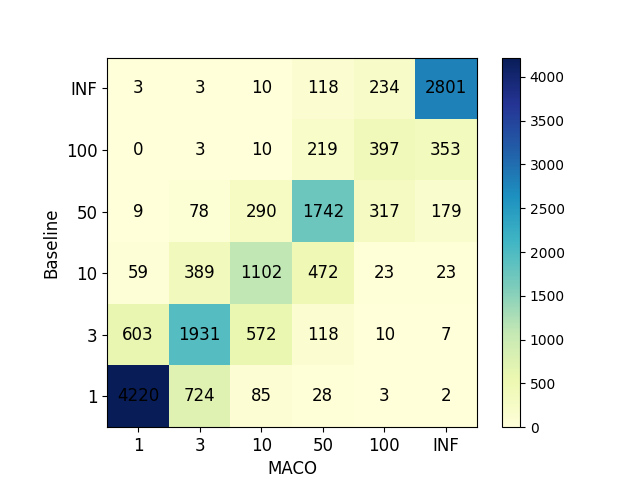}}
  \subfigure[Modality-complete triples]{\includegraphics[width=0.45\linewidth]{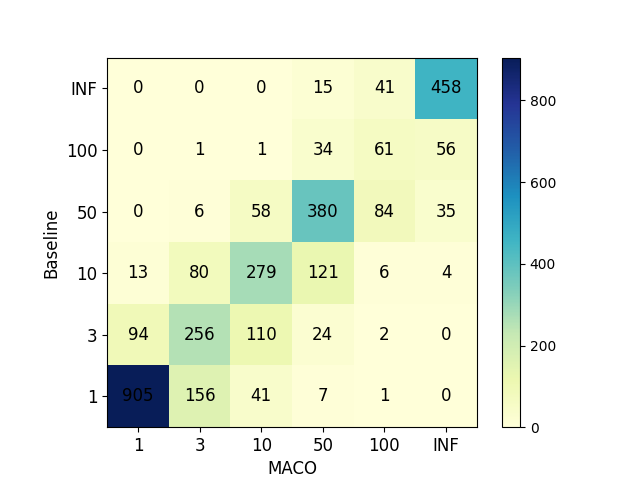}}
  
  \caption{Heat map visualization for the link prediction results. In each heat map, the x-axis/y-axis represents the link prediction results enhanced by MACO and random completion respectively. We divided the linked prediction results into six intervals.}
  \label{heat}
\end{figure}
To illustrate the effectiveness of MACO and answer \textbf{RQ3}, we further conduct a case study. We divide the triples in the test set into two categories based on whether or not there is a modality-missing entity in the triple. Besides, we draw the heat maps of the link prediction results between the MACO-enhanced model and the baseline model. The heat maps are shown in Figure \ref{heat}, where the lower half of the diagonal indicates the triples where the MACO method outperformed the baseline model, and the upper half indicates the opposite. 
\par We find though some triples get worse rankings, more modality-missing triples achieve better ranks with the help of MACO, which reflects that MACO could complete the missing visual information with semantic-rich visual representations. For example, given the test triple \textit{(Michael Gough, /film/actor, Batman)}, the tail entity \textit{Batman} is modality-missing. Typically, the visual information of the film might be a poster which is important information to match the actors, which is similar to the case mentioned in Figure \ref{case}. Thus, the modality-missing situation makes the predicted rank of the baseline model 60. However, the model enhanced by MACO predicts the correct tail entity with rank 1. Such a simple case intuitively demonstrates the effectiveness of MACO. Besides, we could conclude that All-Gen could also benefit those modality-complete triples by generating high-quality visual representations and replacing the original ones.

\section{Conclusion}
In this paper, we mainly discuss the modality-missing problem in the existing MMKGC methods. We argue that vanilla approaches like random initialization would introduce noise into the MMKGC model, leading to bad performance. We propose MACO, a modality adversarial and contrastive framework that generates visual modal features of entities conditioned on structural information to preserve the correspondence between the structure and visual information. This approach completes modality-missing entities with semantic-rich visual representations. We conduct experiments on public benchmarks to demonstrate the effectiveness of MACO. In the future, we plan to collaborative the collaborative design of missing-modality completion and knowledge graph completion.

\section*{Acknowledgement}
This work is funded by Zhejiang Provincial Natural Science Foundation of China (No. LQ23F020017), Yongjiang Talent Introduction Programme (2022A-238-G), and NSFC91846204/U19B2027.

%
%
%

\bibliography{ref.bib}
\bibliographystyle{splncs04}

\end{document}